\documentclass{article}
\usepackage[utf8]{inputenc}
\usepackage{amsmath, amssymb}
\usepackage{hyperref}
\usepackage{graphicx}
\usepackage{float}
\usepackage[T1]{fontenc}
\usepackage[french]{babel}
\usepackage{csquotes}
\usepackage[utf8]{inputenc}
\usepackage[table]{xcolor}
\usepackage{caption}
\usepackage{array}        
\usepackage{multirow}     
\usepackage{colortbl}     
\usepackage{array}
\usepackage{etoolbox}
\usepackage{authblk}
\usepackage{bbm}

\AtBeginEnvironment{tabular}{\centering}
\setkeys{Gin}{keepaspectratio=true}
\usepackage{enumitem}
\setlist[itemize]{label=\textbullet}

\usepackage[T1]{fontenc} 
\usepackage{lmodern} 
\usepackage{microtype} 

\usepackage[shortcuts]{extdash} 
\usepackage{hyphenat} 

\addto\captionsfrench{}
\addto\captionsfrench{\renewcommand{\abstractname}{Abstract}}
\title{\textbf{Intra-neuronal attention within language models \\
\large Relationships between activation and semantics}}

\author{Michael Pichat$^{1,2,4}$, William Pogrund$^{1,5}$, Paloma Pichat$^{1,3}$, Armanouche Gasparian$^{1}$, Samuel Demarchi$^{1,4}$, Martin Corbet$^{1,2}$, Alois Georgeon$^{1,2}$, Michael Veillet-Guillem}

\date{}

\affil[1]{Neocognition (Chrysippe R\&D)}
\affil[2]{Facultés Libres de Philosophie et de Psychologie de Paris (ER IPC)}
\affil[3]{Faculté de Médecine de Lyon Est (Université Lyon 1)}
\affil[4]{Université Paris 8}
\affil[5]{INP-PHELMA, Université Grenoble Alpes}
\begin{document}

\maketitle

\renewcommand{\abstractname}{Abstract}
\begin{abstract}
This study investigates the ability of perceptron-type neurons in language models to perform intra-neuronal attention; that is, to identify different homogeneous categorical segments within the synthetic thought category they encode, based on a segmentation of specific activation zones for the tokens to which they are particularly responsive. The objective of this work is therefore to determine to what extent formal neurons can establish a homomorphic relationship between activation-based and categorical segmentations. The results suggest the existence of such a relationship, albeit tenuous, only at the level of tokens with very high activation levels. This intra-neuronal attention subsequently enables categorical restructuring processes at the level of neurons in the following layer, thereby contributing to the progressive formation of high-level categorical abstractions.

\end{abstract}

\section{Theoretical Context}

\subsection{Attention Heads}

Before the rise of transformers, natural language processing models primarily relied on recurrent (RNN, LSTM) and convolutional (CNN) architectures. While RNNs and LSTMs could capture long-term dependencies through their memory mechanisms, they suffered from the vanishing gradient problem, making it difficult to learn relationships over long sequences. Moreover, their sequential nature limited parallelization and slowed down training. CNNs, on the other hand, offered better parallelization but were ill-suited for global dependencies due to their limited receptive field, which depended on network depth. These approaches thus struggled with capturing long-range dependencies and faced parallelization challenges \cite{Bahdanau2014, Cho2014}. These limitations motivated the introduction of transformers and the self-attention mechanism, which overcame these constraints by enabling efficient parallel processing while capturing long-range relationships. The introduction of self-attention by Vaswani \cite{Vaswani2017} marked a major turning point in NLP.

Transformers leverage self-attention, where each element in a sequence weighs the importance of other elements, facilitating the modeling of complex dependencies. Multi-head attention enhances this approach by allowing different attention heads to specialize in various aspects of representation. Devlin et al. \cite{Devlin2019} demonstrated that certain BERT heads capture syntactic relations, such as subject-verb links, while others focus on more global semantic relationships. Radford et al. \cite{Radford2018} showed that multiple attention heads in GPT improve sentence context modeling by capturing distributed information across different input sequence positions. This capability enables transformers to enhance the richness and hierarchy of representations, improving generalization across various tasks.

Within a transformer, each attention layer comprises multiple heads, each performing an attention operation on a linear projection of the inputs. Self-attention projects representations into three sets of vectors: queries (Q), keys (K), and values (V). For an input sequence \( X \in \mathbb{R}^{T \times b} \), where \( T \) is the sequence length and \( b \) the vector dimension, these matrices are defined as:

\[
Q = X W^T, \quad K = X W^T, \; V = X W^T
\]

where \( W^T \in \mathbb{R}^{b \times b_k} \) are learned weight matrices, and \( b_k \) is the dimension of keys and queries. Attention is computed using the scaled dot-product attention:

\[
\text{Attention}(Q, K, V) = \text{softmax} \left( \frac{Q K^T}{\sqrt{d_k}} \right) V
\]

The normalization by \( \sqrt{d_k} \) stabilizes training by preventing large-scale values in dot products \cite{Vaswani2017}. Multi-head attention applies these operations in parallel over \( h \) different projections of \( Q, K, \) and \( V \):

\[
\text{head}_i = \text{Attention}(Q W_i^G, K W_i^K, V W_i^V)
\]

The concatenation of heads followed by a linear projection yields the final output:

\[
\text{MultiHead}(Q, K, V) = \text{Concat}(\text{head}_1, \dots, \text{head}_h) W^O
\]

This architecture enables the extraction of contextual information from multiple perspectives and enhances the capture of complex relationships in data.

Studies on attention mechanisms have shown that some heads are specialized, while others are redundant. Luong et al. \cite{Luong2015} demonstrated that certain heads capture precise syntactic relations, whereas others focus on global semantic relationships. \cite{Voita2019, Michel2019} observed that removing several heads does not significantly impact model performance, suggesting compensatory mechanisms among the remaining heads.

Statistical mechanics approaches have been used to analyze interactions between attention paths. \cite{Tiberi2024} modeled the contribution of attention heads via a kernel decomposition:

\[
K = \sum_{i=1}^{h} K_i
\]

Each kernel \( K_i \) corresponds to a specific head, allowing an evaluation of its role in the model’s final representation. Results indicate that some heads play a structuring role, while others can be eliminated without significant impact. This observation paves the way for transformer architecture optimizations by reducing redundant heads and improving model interpretability.

The computational efficiency of transformers has been extensively researched. Sparse Transformers \cite{Child2019} reduce attention complexity to \( O(n \log n) \) by introducing a sparse attention structure. Reformer \cite{Kitaev2020} optimizes memory management via key-value factorization and local attention. Performer \cite{Zaheer2020} replaces standard attention with a linear approximation, reducing complexity to \( O(n) \). This method relies on random projections of keys and queries into a lower-dimensional space, where dot products are computed approximately using kernels favoring efficient factorization. This avoids costly dense matrix multiplications while maintaining high accuracy, making attention scalable even for long sequences. Longformer and BigBird \cite{Tay2022} combine local and global attentions to efficiently process long sequences.

Other studies have analyzed attention head specialization in specific contexts. Clark et al. \cite{Clark2019} examined BERT attention matrices and found that some heads learn specific syntactic relationships, such as subject-verb dependencies or anaphoric relations. Transformer-XL \cite{Dai2019} introduced a recurrent memory mechanism that captures longer-term dependencies, improving text generation and dialogue modeling.

Finally, attention mechanisms have extended to other domains, including computer vision with Vision Transformer (ViT) \cite{Dosovitskiy2021} and Swin Transformer \cite{Liu2021}, as well as neuroscience and cognitive process modeling \cite{Traylor2024}.

\subsection{Human Attention}

Biologically, human attention arises from the limited capacity of the neuronal system to process information. It manifests through selective mechanisms in the acquisition, activation, and utilization of sensory or memory data (such as knowledge or rules) \cite{Funayama2024, Barr2024}. This results in an oriented response, focusing information retrieval on specific characteristics. Neurocognitive research on attentional mechanisms, inspired notably by Posner's work \cite{Posner1995, Posner2024, Rueda2024}, highlights the existence of a frontal attention system associated with conscious attention and planning, and a posterior system, located in the parietal lobe, involved in visuospatial processes and shifts in attentional focus.

In cognitive psychology, attention is conceptualized as the calibration of activity toward a specific goal, thereby enhancing efficiency in the collection and execution of information (selectivity, accuracy, speed) for a given task \cite{Posner1975, Schneider1977, Posner1978, Richard1980, Treisman1980, Duncan1984, Tipper1985, Cowan2024, Wu2024, Gresch2024}. During task execution, attention is managed by the central nervous system, which determines the relevance of internal information (such as knowledge or schemas) that ensures execution quality.

Generally, two cognitive functions are associated with attention \cite{Duncan1999}:
\begin{itemize}
    \item Signal detection, which primarily relies on vigilance and exploration to identify the appearance of a specific stimulus.
    \item Selective attention, focused on specific stimuli while excluding others.
\end{itemize}

Vigilance refers to the ability to concentrate on a flow of information to detect a specific signal \cite{Mackworth1948}, which may appear rarely but requires a rapid reaction \cite{Chen2024, Hanzal2024}. It is negatively affected by the level of uncertainty regarding the targeted elements \cite{Broadbent1965}. Vigilance can be defined as an adjustable attentional beam influenced by the anticipation of the signal’s appearance at a specific location \cite{Posner1980, Motter1999, Murray2024}.

\textit{Visual exploration} \cite{Zhang2024, Wu2024}, on the other hand, involves an active search for stimuli, contrasting with the passive expectation of their emergence \cite{Posner1998}. It is characterized by a scanning strategy for recognizing attributes within a given environment. According to the \textit{Feature Integration Theory} \cite{Treisman1986, Rosenholtz2024}, a mental map for each visual attribute represents occurrences within the visual field, which are regularly inspected in parallel. Here, attentional processes play a role in mental binding, assembling various attributes of the same object while inhibiting irrelevant characteristics. The \textit{Similarity Theory} \cite{Duncan1992} analyzes attentional exploration as an evaluation of the proximity between target stimuli and distractors. Meanwhile, the \textit{Guided Search Theory} \cite{Cave1990, Alahmari2024} divides exploratory attention into two phases: first, the activation of a global representation of potential targets, followed by a serial analysis to identify the most activated target.

Selective attention is often explored through the \textit{cocktail party effect} \cite{Cherry1953, Liu2024, Bosker2024}, referring to the ability to follow a conversation among others in the background. The characteristics of this attentional focus include sensory properties, sound volume, and the spatial location of the target speech. In this context, the \textit{Filter Theory} \cite{Broadbent1958, Zhao2024} suggests that a filter selects sensory streams to receive deep processing. However, this model evolved into an \textit{attenuation approach} \cite{Treisman1964, Lin2024}, where all information is reduced in perceptible intensity, leaving only residual data close to targeted criteria. The distribution of limited attentional resources \cite{Kahneman1973} is also relevant for managing multiple tasks in parallel with increased efficiency.

The aspects of attentional mechanisms relevant to our study lie at the intersection of vigilance and selective attention. These processes—detecting a targeted type of information and selectively focusing on specific data characteristics—correspond to the phenomenon of interest here: the impact of a neuron's activation level in response to an incoming token on the intra-neuronal detection and selective attention of tokens exhibiting specific categorical features.

\subsection{Attentional Processes and Conceptualization}

The notion of conceptualization \footnote{Traditionally, research on concepts considers them as cognitive units stored in memory and linked to a word or expression, with an associated class of objects sharing common properties. Experimentally, these approaches have led to studies on familiarity or typicality judgments, category comparisons, classification, and more generally, categorical identification—studies that we have previously synthesized \cite{Pichat2024a, Pichat2024b}. However, the term \textit{conceptualization}, as used by Vergnaud \cite{Vergnaud2016}, does not refer to these classical approaches, which focus on the predicative form of the concept. Instead, it aligns with a developmental and pragmatic perspective, emphasizing the primary form of the concept: its operational form.}is one of the major contributions of the theory of conceptual fields proposed by Vergnaud \cite{Vergnaud2009, Vergnaud2016}. As we will see below, conceptualization is a central cognitive process at the intersection of the vigilance and selective attention mechanisms mentioned earlier, specifically in the context of categorizing information received by a cognitive system. This notion, developed in the field of human thought, is therefore particularly relevant, heuristically, to our current investigation concerning the identification and selective attention of specific data (tokens) at the level of internal processing carried out by formal neurons.

Conceptualization is an attentional representational activity whose purpose is the identification of the operational characteristics of stimuli (in our case, tokens) to which it applies, in order to ground a cognitive system’s activity on these characteristics and thus enhance its efficiency. The cognitive function of conceptualization is to extract an operational form of knowledge, which becomes the object of specific attentional focus. In the domain of human cognition, this attentional focus is largely unconscious and non-verbalizable; for this reason, Vergnaud \cite{Vergnaud2009} refers to it as knowledge-in-action.

Vergnaud \cite{Vergnaud2016} defines a concept-in-action, resulting from the selective attentional activity of conceptualization, as a category of thought learned to be relevant for a given task (in our case, the targeted processing of textual data). Regarding this notion of concept-in-action, three key points should be highlighted:
\begin{enumerate}
    \item Concepts-in-action are categories of thought through which a cognitive system identifies, selects, and captures certain pieces of information present in a given situation (a set of tokens in the case of language processing). In other words, concepts-in-action function as cognitive attentional filters that allow a given situation to be selectively ``read'' or ``perceived.''
    \item From an epistemological perspective, there exists a potentially infinite number of formal types of thought categories. The most commonly encountered types are: object, property, relation, transformation, condition, and process.
    \item Concepts-in-action serve as pragmatic vectors of synthetic thought, organizing attentional information processing by segmenting the token space according to the contingent goals of the finalized activity for which a language model has been trained. Indeed, the functionality of concepts-in-action lies in their ability to enable the neural system to focus its attention on a limited number of selected elements, learned to be crucial for the success of the synthetic system’s activity. As such, they support a representation of only those situational variables whose consideration is essential for the effectiveness of the task at hand.
\end{enumerate}

The approach developed by Vergnaud \cite{Vergnaud2016} establishes conceptualization as a fundamentally economic and pragmatic attentional cognitive activity. This inherently pragmatic finalization of conceptualization is the reason why the concepts it extracts are \textit{in action}, meaning they are encapsulated within the cognitive system’s activity.

An artificial neuron can be described as a synthetic cognitive operator of conceptualization, whose purpose is to select, from a set of incoming tokens, one or more specific subgroups of tokens. These subgroups constitute the categorical extension of the \textit{critical} concept(s)-in-action that the neuron is functionally tasked with identifying selectively. In this sense, the neuron performs an attentional focusing activity on certain types of tokens that need to be selected and filtered to optimize the efficiency of the linguistic processing task in which the neuron participates.

\section{Research Problem}

\subsection{Mathematico-Cognitive Factors of Categorical Segmentation and Intra-Neuronal Attention}

In a previous study \cite{Pichat2024c}, we explored the mathematico-cognitive factors influencing how an artificial neural network (of the perceptron type) in a language model performs categorical segmentation of the tokens presented to it. Based on the neuronal aggregation function of the form \( \sum (w_{i,j} \; x_{i,j} ) + b \), which partially governs this cognitive process, we identified three factors contributing to this conceptual partitioning:

\begin{itemize}
    \item The \( x \)-effect: Synthetic Categorical Priming. This effect refers to the influence of synthetic thought categories activated in neurons of layer \( n \) on the activation of categories in neurons that are strongly connected via attention in the next layer. In other words, a token that strongly belongs to an initial category in layer \( n \) is more likely to belong to a strongly associated category in layer \( n+1 \).
    
    \item The \( w \)-effect: Synthetic Inter-Categorical Attention. This effect influences the degree of importance that a receiving neuron (in layer \( n+1 \)) assigns to the categories of neurons in the preceding layer (\( n \)), based on connection weights. This process results in categorical complementation, where each attentively focused precursor category contributes a unique sub-dimensional category (comprising highly specific tokens) to the formation of the receiving category. Thus, the category of a receiving neuron is constructed by assembling complementary categorical sub-dimensions derived from its precursor categories.
    
    \item The \( \sum \)-effect: Synthetic Categorical Phasing. This effect refers to the tendency for tokens that are jointly activated within precursor categories (layer \( n \)) to be more likely to be part of the categorical extension of their strongly associated attentionally focused category in layer \( n+1 \). This process manifests as categorical intersection.
\end{itemize}

The phenomenon of interest here, as a reminder, is the extent to which the specific activation level of a neuron in response to a token is linked to a particular categorical value of that token within the category carried by the neuron. In other words, is a given segment of activation values associated with a distinct categorical segment, allowing a neuron to selectively focus its attention on a specific categorical segment based on its associated activation span?

The three mathematico-cognitive factors we have just outlined—which could be related to the genesis and structuring of human thought categories \cite{Bolognesi2020, Haslam2020, Eysenck2020, Fel2024, Bathia2024, Marconato2024, Zettersten2024}—are particularly relevant to our research question. Indeed, these three factors directly determine, before the application of the nonlinear activation function, the activation value that the neuronal aggregation function assigns to a given token. In other words, these factors define the activation zone, the activation segment within which a processed token is positioned by the involved neuron.

To what extent do these three factors—both as a matrix of activation values assigned to tokens by a neuron and as carriers of categorical effects, as previously described—delineate specific activation segments paired with specific categorical segments that could thus become the object of selective attentional focus? Alternatively stated, how are these factors, through neuronal activation values, associated with the conceptualization of certain categorical segments, on which it is both relevant and effective for the neuron to focus its internal attention?

\subsection{Synthetic Categorical clipping and Intra-Neuronal Attention}

In a previous study \cite{Pichat2025}, we highlighted that the three mathematico-cognitive factors of categorical segmentation, as detailed above, govern a mechanism of \textit{synthetic categorical clipping}. This clipping process results in the elaboration and separation of a form from a categorical background. More precisely, categorical clipping is the phenomenon in synthetic cognition by which a specific categorical sub-dimension is extracted from the category carried by a precursor neuron (in layer \( n \)) to contribute to the formation of a superordinate category (in layer \( n+1 \)).

Categorical clipping manifests through a series of synthetic characteristics:
\begin{itemize}
    \item Categorical reduction, meaning that the categorical sub-dimension extracted from a precursor category contains tokens that are semantically more homogeneous compared to the original category.
    \item Categorical selectivity, referring to the extraction of a small subgroup of tokens from the larger set of tokens that initially characterized the original category.
    \item Separation of initial embedding dimensions, associated with a differentiation of these embeddings, with some being more specifically related to the outlined categorical sub-dimension.
    \item Partitioning of categorical zones within initial embedding dimension, where certain zones are more specifically associated with the extracted sub-dimensions.
\end{itemize}

Categorical clipping is an activity that extracts a categorical sub-dimension from the category associated with a neuron in layer \( n \), but this process occurs \textit{externally} to the originating neuron, specifically within one of its paired neurons in layer \( n+1 \). However, to what extent is this clipping—ultimately driven by activation values—executed based on intra-neuronal attentional focus and conceptualization of specific categorical sub-segments within the original category, sub-segments that would be associated with particular activation segments?

\subsection{Categorical Restructuring and Intra-Neuronal Attention}

In a previous study \cite{Pichat2025b}, we explored the process of synthetic categorical restructuring, specifically the generation, at each neuronal layer \(n+1\), of new artificial thought categories that are more functional for segmenting the world of tokens in alignment with the purpose of the neural network's activity. This process falls under reflective abstraction in the Piagetian sense \cite{Pichat2025}, applied to the categories of layer \(n\).

This categorical restructuring is directly dependent on the coactivity of the three factors of categorical segmentation outlined earlier (categorical priming, inter-categorical attention, and categorical phasing). We posited that the restructuring phenomenon is particularly linked to inter-categorical attention, given that this latter factor, by mathematical construction of the neuronal aggregation function, is both a necessary condition and an amplifier of the other two factors (priming and phasing).

We demonstrated that the joint action of inter-categorical attention and categorical phasing leads to partial categorical confluence: the categorical sub-dimensions extracted from layer \(n\) categories, at the level of a neuron that is strongly attentive to them in layer \(n+1\), tend to semantically converge to a relative extent. We also highlighted that the combined impact of inter-categorical attention and categorical priming generates activation dispersion: a categorical sub-dimension extracted from a category in layer \(n\) does not correspond to a continuous segment of token activations at the originating neuron.

How are both this partial categorical confluence and this activation dispersion, occurring during the transition from layer \(n\) to layer \(n+1\), realized through conceptualization and intra-neuronal attentional focus at the level of layer \(n\)? In other words, how does partial categorical confluence, driven by the coactivity of synthetic factors such as inter-categorical attention and categorical phasing, become possible through the identification—via activation—of specific categorical sub-segments to be considered at the level of the involved precursor neuron? And to what extent is activation dispersion compatible with intra-neuronal attentional focus based on specific activation segments, which might appear paradoxical?

\subsection{Conceptualization and Intra-Neuronal Attention}

From a mathematical perspective, a synthetic neuronal processing unit results from the composition of functions: a nonlinear activation function (ReLU, SELU, GELU, ELU, etc.) applied to a linear aggregation function of the form \( \sum (w_{i,j} \; x_{i,j}) + b \). Within a language model, this matrix-based processing associates an input token (or rather its embedding) with an output activation value. 

What is the epistemological significance of this activation value from a semantic standpoint? Is this activation value correlated with a specific semantic value? Are different activation segments within the activation space associated with distinct categorical segments in the semantic space of the involved neuron? For a given neuron, is there a homomorphic relationship between its activation space and its categorical space? Can these spaces be divided and segmented into identifiable and pairable activation and categorical zones? 

In other words, is it possible to locate intra-neuronal activation segments that correspond to specific sub-semantic structures within the thought categories represented by the neurons? Put differently, to what extent does the activation value serve as a quantifier enabling conceptualization—an attentional mapping of specific intra-neuronal semantic zones?

\section{Methodology}

\subsection{Methodological Positioning}

To situate our current research methodologically, we position it within a set of explainability investigation techniques aimed at making artificial neural networks more comprehensible. These methods seek, with varying degrees of cognitive depth, to explain internal mechanisms or interpret the meaning and function of information flows within these networks, whether studied at the level of individual neural layers, groups of layers, or the entire model.

Studies focusing on \textit{macroscopic explainability} examine fluctuations between input data and outputs to clarify the relationship between what is given to the system and what it produces. In this approach, gradient-based methods assess the influence of each input by analyzing the partial derivatives of each input dimension \cite{Enguehard2023}. Input attributes can be evaluated through various elements, including feature importance \cite{Danilevsky2020}, token relevance \cite{Enguehard2023}, or attention coefficients \cite{Barkan2021}. Furthermore, example-based methods observe variations in outputs when input modifications are introduced, allowing for an analysis of how slight data alterations affect model predictions \cite{Wang2022}, as well as assessing the implications of input perturbations such as deletion, negation, mixing, or masking \cite{Atanasova2020, Wu2020, Treviso2023}. Other approaches focus on conceptually mapping inputs to evaluate their influence on observed outputs \cite{Captum2022}.

\textit{Microscopic explainability} methods, on the other hand, examine the intermediate internal states of language models rather than their overall output, analyzing interconnections and activations of individual neurons or groups of neurons. Some studies explore how to segment and interpret neural activations in a layer based on the inputs from the previous layer \cite{Voita2021}. Others aim to adjust activation functions to enhance interpretability \cite{Wang2022}. Certain techniques investigate the knowledge embedded within neurons, extrapolating internal meanings through significance matrices \cite{Dar2023, Geva2023}. Finally, some approaches assess statistical patterns in neural responses based on specific datasets \cite{Bills2023, Mousi2023, Durrani2022, Wang2022, Dai2022}. Our current work falls within this latter methodological category.

\subsection{Model and Statistical Units Used}

Building on our previous work \cite{Pichat2024a, Pichat2024b, Pichat2024c, Pichat2024d, Pichat2025}, we focused on OpenAI’s transformer-based GPT model, specifically the GPT-2XL version. This choice was made because GPT-2XL offers sufficient complexity to investigate advanced cognitive processes while avoiding, for an initial exploration, the interpretative challenges posed by GPT-4 or GPT-4o. In 2023, OpenAI released detailed documentation on the parameters of GPT-2XL, as outlined by Bills et al., \cite{Bills2023}, which we leveraged in our current investigation.

To reduce the scope of our statistical analyses, we focused on the first two perceptron layers of GPT-2XL, each containing 6,400 neurons, for a total of 12800 artificial neurons. Regarding linguistic elements (tokens) and their associated activations, our study centered on the 100 tokens exhibiting the highest mean activations per neuron, which we termed \textit{core-tokens}. When analyzing relationships between neurons in layers 0 and 1, we limited our selection for each neuron in layer 1 to its 10 most strongly connected neurons from layer 0, based on attention weight values.

\subsection{Statistical Processing}

For our statistical analyses, we used the SciPy library, in accordance with the recommendations of \cite{Howell2024, Beaufils1996, Ellis2010, Ellis2020}.

Our verification of the normality of the data, to assess the feasibility of parametric tests, was conducted in two steps. First, we applied inferential tests: the Shapiro-Wilk test, suitable for small datasets; the Lilliefors test, used when distribution parameters are unknown; the Kolmogorov-Smirnov test for large samples; and the Jarque-Bera test, which quantifies skewness and kurtosis in large samples. Complementing this, in a second step, we examined skewness and kurtosis coefficients and visualized distributions using QQ-plots to compare the recorded data with a theoretical normal distribution. To check the homogeneity of variances among data subgroups whose relationships were being analyzed, we applied Bartlett's test, which is sensitive to deviations from normality, and Levene's test, which is less affected by such deviations.

These preliminary checks indicated limited normality in our data. Consequently, our statistical analyses relied on non-parametric approaches, specifically:

\begin{itemize}
    \item The Kruskal-Wallis test, used to investigate the relationship between a categorical variable defining independent groups and an ordinal variable. This test was applied based on ranking the numerical activation data of tokens, adhering to the standard conditions for its application, including groups with a minimum of six observations. Effect sizes for the Kruskal-Wallis test were measured using Cohen’s \( d \). For pairwise group comparisons, given \( k \) total groups, \( k(k-1)/2 \) comparisons were performed, using a rank difference coefficient adapted for post hoc situations, measured on a \( z \)-scale, with a significance threshold \( \alpha \) divided by \( k(k-1) \).
    
    \item The univariate \(\chi^2\) goodness-of-fit test, applied while ensuring compliance with its theoretical and observed frequency requirements, thus avoiding alternatives suited for small samples, such as Fisher’s or Monte Carlo methods. Effect sizes were measured using the risk ratio.
\end{itemize}

For certain analyses, we employed a typological classification approach via hierarchical clustering. This classification was configured as follows:  
(i) a top-down classification approach,  
(ii) the use of Euclidean distance to measure distances between statistical units,  
(iii) a predefined number of five clusters,  
(iv) Ward’s method as the aggregation technique,  
(v) prior standardization of the data.

\subsection{Methodological Operationalizations}

To quantify semantic proximity between tokens, we used cosine similarity based on the embedding space of GPT-2XL. This approach allowed us to avoid methodological limitations identified by Bills et al., \cite{Bills2023} when attempting to relate various artificial cognitive systems using non-unified embedding foundations.

To examine the potential relationship, for each given neuron, between its categorical segments and its activation segments, we employed two opposing methodological approaches:

\begin{itemize}
    \item A "top-down" approach, which starts from predefined categorical segments and investigates the extent to which they are associated with distinct activation segments. These categorical segments were first defined using hierarchical clustering based on the GPT-2XL embeddings of the 100 \textit{core-tokens} defining the category associated with each investigated neuron. Then, for methodological diversification, segmentation was also performed using prompt engineering with OpenAI's GPT-4o model. This first approach was further extended by measuring the interweaving and activation overlaps of the categorical token segments obtained via prompt engineering.
    
    \item A "bottom-up" approach, which instead starts from predefined activation segments and evaluates whether these segments exhibit semantic proximity. To ensure methodological diversification, activation segments were defined first through activation quartiles and then through hierarchical clustering of activations.
\end{itemize}

\section{Results}

As a reminder, our research question is as follows: does a synthetic process of conceptualization and intra-neuronal attention exist, enabling each neuron to identify and isolate specific categorical segments within the artificial thought category it carries, based on determined activation segments? 

We operationalized our empirical investigation of this question through two methodologically opposing studies, which we will present sequentially:

\begin{itemize}
    \item A "top-down" study, in which we started from categorical segments to examine their respective average activation values.
    \item A "bottom-up" study, in which we first isolated specific activation segments and then analyzed their respective categorical homogeneity.
\end{itemize}

\subsection{Activation Differentiation of Categorical Clusters}

As part of our \textit{top-down} study, for each perceptron neuron in layers 0 and 1 of GPT-2XL, we decomposed the thought category it carries (through its 100 \textit{core-tokens}, i.e., the tokens with the highest average activation) into five categorical clusters. These clusters represent subcategories that are relatively homogeneous according to a given semantic criterion, allowing the initial category to be partitioned meaningfully.

To illustrate this, we present below the example of neurons n°5 (layer 1) and n°5065 (layer 0) using our \href{https://neuron-viewer.neocognition.org/1/5}{\textit{genetic neural viewer}}. This serves to investigate the relationship between categorical segmentation (clustering) and activation segmentation. More specifically, we aim to determine whether there is a significant difference in average activation values among the categorical clusters obtained.

\begin{figure}[H]
    \centering
    \includegraphics[height=4.2cm]{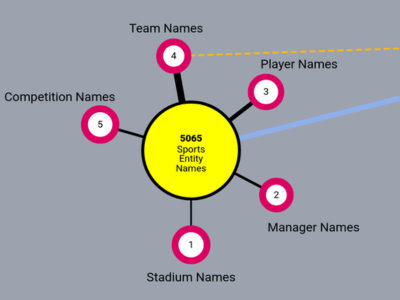}
    \hspace{0.05\textwidth}
    \includegraphics[height=4.2cm]{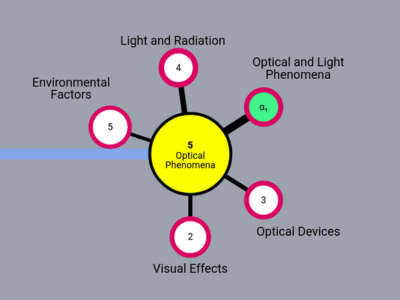}
\end{figure}

A first operationalization of our approach involved generating five categorical token clusters for each neuron, based on hierarchical clustering applied to the GPT-2XL embeddings of the 100 \textit{core-tokens} associated with that neuron. From a methodological perspective, we clarify the following points:

\begin{itemize}
    \item The use of GPT-2XL embeddings, like any operationalization, is a specific methodological choice. It does not claim to measure any intrinsic semantic reality—an epistemologically meaningless notion—but rather represents one particular modality, among many possible, for evaluating semantic phenomena within a given contingent semantic space. Our interpretations must therefore remain confined to this specific semantic contingency.
    
    \item In our overall statistics, we retained only those neurons whose semantic clustering resulted in clusters containing at least six tokens, ensuring compliance with the conditions for using the Kruskal-Wallis inferential test for mean comparisons.
    
    \item For each neuron, we systematically designated as "\( K_1 \)" the categorical cluster with the lowest mean activation value among its constituent tokens, and so forth up to "\( K_5 \)."
\end{itemize}

Table n°1 summarizes our results for the 2194 neurons analyzed in layer 0. Overall, we observe a low percentage (21.46\%) of neurons exhibiting a significant difference (\(\alpha = 5\%\)) in mean activations (\(\mu_{K_n}\)) across their five categorical clusters. This finding is corroborated by the low percentages (\(\pi(p_{K_n,K_m} < \alpha')\)) of significant mean activation differences in post hoc pairwise comparisons of categorical clusters (with an adjusted significance threshold \(\alpha' = \alpha / 20\)). 

However, an interesting trend emerges: while the mean activation distances (\(\mu(\delta_{K_n,K_m})\)) between successive categorical clusters are generally small, the average activation distance between clusters \( K_4 \) and \( K_5 \) (\(\mu(\delta_{K_4,K_5}) = .0955\)) is slightly larger than that between clusters \( K_1 \) and \( K_2 \) (\(\mu(\delta_{K_1,K_2}) = .0512\)). This trend aligns with the slightly higher effect size (measured using Cohen's \( d \)) for the activation distance between \( K_4 \) and \( K_5 \) (\(\mu(d_{K_4,K_5}) = .2785\)) compared to \( K_1 \) and \( K_2 \) (\(\mu(d_{K_1,K_2}) = .2404\)). 

More strikingly, this trend becomes much more pronounced when considering the strong effect size of the mean activation difference between clusters \( K_1 \) and \( K_5 \) (\(\mu(d_{K_1,K_5}) = .8202\)). One might hypothesize that the weak significance of post hoc inferential tests comparing \(\mu_{K_1}\) and \(\mu_{K_5}\) (\(\pi(p_{K_1,K_5} < \alpha') = 14.02\%\)) is due to a bias resulting from the small number of involved tokens. 

Graph 1 visually summarizes the main data mentioned here.

\begin{table}[H]
    \centering
    \renewcommand{\arraystretch}{1.3}
    \begin{tabular}{|>{\columncolor{gray!20}}c|>{\columncolor{gray!20}}c|>{\columncolor{gray!10}}c|>{\columncolor{gray!10}}c|}
        \hline
        \multicolumn{2}{|c|}{{N\textsubscript{neuron}}} & \multicolumn{2}{c|}{2194} \\
        \hline
        \multicolumn{2}{|c|}{$\pi(p_{\text{KW}}<.05)$} & \multicolumn{2}{c|}{21.4612} \\
        \hline
        $\mu_{K_1}$ & 1.7117 & $\mu(\delta_{K_1,K_2})$ & .0512 \\
        $\mu_{K_2}$ & 1.7629 & $\mu(\delta_{K_2,K_3})$ & .0416 \\
        $\mu_{K_3}$ & 1.8045 & $\mu(\delta_{K_3,K_4})$ & .0484 \\
        $\mu_{K_4}$ & 1.8530 & $\mu(\delta_{K_4,K_5})$ & .0955 \\
        $\mu_{K_5}$ & 1.9485 & $\mu(\delta_{K_1,K_5})$ & .2368 \\
        \hline
        \rowcolor{gray!30} $\mu(d_{K_1,K_2})$ & .2404 & \cellcolor{gray!40} $\pi(p_{K_1,K_2}<\alpha')$ & \cellcolor{gray!40} .1370 \\
        \rowcolor{gray!30} $\mu(d_{K_2,K_3})$ & .1650 & \cellcolor{gray!40} $\pi(p_{K_2,K_3}<\alpha')$ & \cellcolor{gray!40} .0913 \\
        \rowcolor{gray!30} $\mu(d_{K_3,K_4})$ & .1664 & \cellcolor{gray!40} $\pi(p_{K_3,K_4}<\alpha')$ & \cellcolor{gray!40} .2283 \\
        \rowcolor{gray!30} $\mu(d_{K_4,K_5})$ & .2785 & \cellcolor{gray!40} $\pi(p_{K_4,K_5}<\alpha')$ & \cellcolor{gray!40} 1.3242 \\
        \rowcolor{gray!30} $\mu(d_{K_1,K_5})$ & .8202 & \cellcolor{gray!40} $\pi(p_{K_1,K_5}<\alpha')$ & \cellcolor{gray!40} 14.0183 \\
        \hline
    \end{tabular}
    \vspace{0.3cm}
    \captionsetup{justification=centering}
    \caption*{\textit{Table n°1: Comparison of mean activations between categorical clusters from hierarchical classification on tokens' embeddings (layer 0).}}
\end{table}

\begin{figure}[H]
    \centering
    \includegraphics[width=0.6\textwidth]{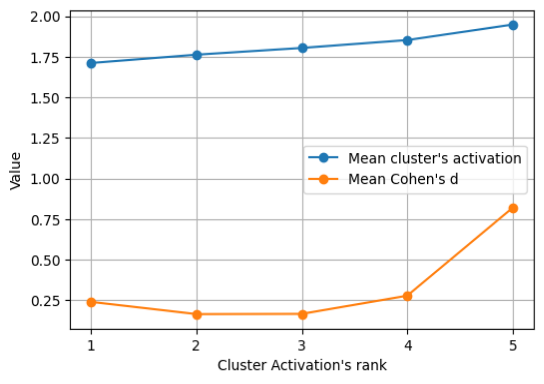}
    \captionsetup{justification=centering, font=small}
    \caption*{\textit{Graph n°1 : Comparison of mean activations between categorical clusters from hierarchical classification on tokens' embeddings (layer 0).}}
\end{figure}

These initial results suggest the following dual trend:

\begin{itemize}
    \item Activation indifferentiation: A weak difference in mean activation values among categorical clusters associated with lower mean activation values.
    \item Relative activation differentiation: A stronger difference in mean activation values among categorical clusters that involve one or more higher mean activation values.
\end{itemize}

Table n°2 and its associated summary graph n°2 display similar results, but with even more pronounced trends. Notably, the effect size contrast is stronger between clusters \( K_4 \) and \( K_5 \) (\(\mu(d_{K_4,K_5}) = .3363\)) compared to \( K_1 \) and \( K_2 \) (\(\mu(d_{K_1,K_2}) = .2631\)). Additionally, when comparing clusters \( K_1 \) and \( K_5 \), we observe an extremely high effect size (\(\mu(d_{K_1,K_5}) = .9962\)), associated with increased significance (\(\pi(p_{K_1,K_5} < \alpha') = 22.08\%\)).

\begin{table}[H]
    \centering
    \renewcommand{\arraystretch}{1.3}
    \begin{tabular}{|>{\columncolor{gray!20}}c|>{\columncolor{gray!20}}c|>{\columncolor{gray!10}}c|>{\columncolor{gray!10}}c|}
        \hline
        \multicolumn{2}{|c|}{N\textsubscript{neuron}} & \multicolumn{2}{c|}{2192} \\
        \hline
        \multicolumn{2}{|c|}{$\pi(p_{\text{KW}}<.05)$} & \multicolumn{2}{c|}{31.4325} \\
        \hline
        $\mu_{K_1}$ & 1.8300 & $\mu(\delta_{K_1,K_2})$ & .1062 \\
        $\mu_{K_2}$ & 1.9362 & $\mu(\delta_{K_2,K_3})$ & .0896 \\
        $\mu_{K_3}$ & 2.0258 & $\mu(\delta_{K_3,K_4})$ & .1119 \\
        $\mu_{K_4}$ & 2.1377 & $\mu(\delta_{K_4,K_5})$ & .2143 \\
        $\mu_{K_5}$ & 2.3519 & $\mu(\delta_{K_1,K_5})$ & .5219 \\
        \hline
        \rowcolor{gray!40} $\mu(d_{K_1,K_2})$ & .2631 & \cellcolor{gray!60} $\pi(p_{K_1,K_2}<\alpha')$ & \cellcolor{gray!60} .0456 \\
        \rowcolor{gray!40} $\mu(d_{K_2,K_3})$ & .1869 & \cellcolor{gray!60} $\pi(p_{K_2,K_3}<\alpha')$ & \cellcolor{gray!60} .3193 \\
        \rowcolor{gray!40} $\mu(d_{K_3,K_4})$ & .2042 & \cellcolor{gray!60} $\pi(p_{K_3,K_4}<\alpha')$ & \cellcolor{gray!60} .5931 \\
        \rowcolor{gray!40} $\mu(d_{K_4,K_5})$ & .3363 & \cellcolor{gray!60} $\pi(p_{K_4,K_5}<\alpha')$ & \cellcolor{gray!60} 2.1898 \\
        \rowcolor{gray!40} $\mu(d_{K_1,K_5})$ & .9962 & \cellcolor{gray!60} $\pi(p_{K_1,K_5}<\alpha')$ & \cellcolor{gray!60} 22.0803 \\
        \hline
    \end{tabular}
    \vspace{0.3cm}
    \captionsetup{justification=centering}
    \caption*{\textit{Table n°2: Comparison of mean activations between categorical clusters from hierarchical classification on tokens' embeddings (layer 1).}}
\end{table}

\begin{figure}[H]
    \centering
    \includegraphics[width=0.6\textwidth]{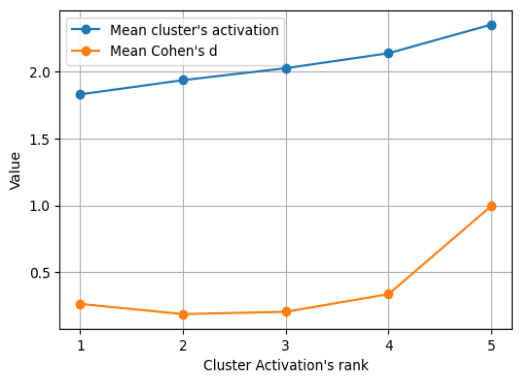}
    \captionsetup{justification=centering, font=small}
    \caption*{\textit{Graph n°2 : Comparison of mean activations between categorical clusters from hierarchical classification on tokens' embeddings (layer 1).}}
\end{figure}

Still within the framework of our \textit{top-down} study, a second operationalization of our approach involved generating five categorical token clusters for each neuron (again based on its 100 \textit{core-tokens}), this time using a system of prompts applied to GPT-4 within a Python-coded structure. This allowed us to investigate our research question from an alternative semantic observation framework coupled with a different clustering methodology.

Table n°3 and its associated summary graph (Graph 3), covering 2316 neurons from layer 0, consistently exhibit the same types of results as previously observed.

\begin{table}[H]
    \centering
    \renewcommand{\arraystretch}{1.3}
    \begin{tabular}{|>{\columncolor{gray!20}}c|>{\columncolor{gray!20}}c|>{\columncolor{gray!10}}c|>{\columncolor{gray!10}}c|}
        \hline
        \multicolumn{2}{|c|}{N\textsubscript{neuron}} & \multicolumn{2}{c|}{2316} \\
        \hline
        \multicolumn{2}{|c|}{$\pi(p_{\text{KW}}<.05)$} & \multicolumn{2}{c|}{18.3938} \\
        \hline
        $\mu_{K_1}$ & 1.7443 & $\mu(\delta_{K_1,K_2})$ & .0629 \\
        $\mu_{K_2}$ & 1.8072 & $\mu(\delta_{K_2,K_3})$ & .0512 \\
        $\mu_{K_3}$ & 1.8584 & $\mu(\delta_{K_3,K_4})$ & .0592 \\
        $\mu_{K_4}$ & 1.9176 & $\mu(\delta_{K_4,K_5})$ & .1015 \\
        $\mu_{K_5}$ & 2.0190 & $\mu(\delta_{K_1,K_5})$ & .2747 \\
        \hline
        \rowcolor{gray!40} $\mu(d_{K_1,K_2})$ & .2981 & \cellcolor{gray!60} $\pi(p_{K_1,K_2}<\alpha')$ & \cellcolor{gray!60} .0864 \\
        \rowcolor{gray!40} $\mu(d_{K_2,K_3})$ & .1982 & \cellcolor{gray!60} $\pi(p_{K_2,K_3}<\alpha')$ & \cellcolor{gray!60} .0432 \\
        \rowcolor{gray!40} $\mu(d_{K_3,K_4})$ & .1982 & \cellcolor{gray!60} $\pi(p_{K_3,K_4}<\alpha')$ & \cellcolor{gray!60} .0000 \\
        \rowcolor{gray!40} $\mu(d_{K_4,K_5})$ & .2803 & \cellcolor{gray!60} $\pi(p_{K_4,K_5}<\alpha')$ & \cellcolor{gray!60} .2591 \\
        \rowcolor{gray!40} $\mu(d_{K_1,K_5})$ & .8985 & \cellcolor{gray!60} $\pi(p_{K_1,K_5}<\alpha')$ & \cellcolor{gray!60} 11.9603 \\
        \hline
    \end{tabular}
    \vspace{0.3cm}
    \captionsetup{justification=centering}
    \caption*{\textit{Table n°3: Comparison of mean activations between categorical clusters from GPT4 clustering prompt on tokens' embeddings (layer 0).}}
\end{table}

\begin{figure}[H]
    \centering
    \includegraphics[width=0.6\textwidth]{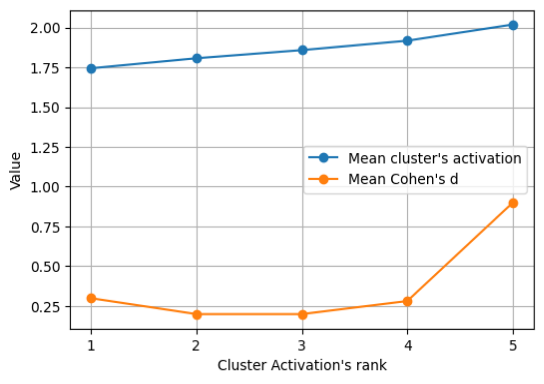}
    \captionsetup{justification=centering, font=small}
    \caption*{\textit{Graph n°3 : Comparison of mean activations between categorical clusters from GPT4 clustering prompt (layer 0).}}
\end{figure}

The same pattern is once again observed in Table n°4 and its corresponding summary graph n°4, which pertain to 1942 neurons from layer 1. This observation could support a hypothesis suggesting that the synthetic phenomena we highlight tend to increase in deeper layers.

\begin{table}[H]
    \centering
    \renewcommand{\arraystretch}{1.3}
    \begin{tabular}{|>{\columncolor{gray!20}}c|>{\columncolor{gray!20}}c|>{\columncolor{gray!10}}c|>{\columncolor{gray!10}}c|}
        \hline
        \multicolumn{2}{|c|}{N\textsubscript{neuron}} & \multicolumn{2}{c|}{1942} \\
        \hline
        \multicolumn{2}{|c|}{$\pi(p_{\text{KW}}<.05)$} & \multicolumn{2}{c|}{29.8661} \\
        \hline
        $\mu_{K_1}$ & 1.8086 & $\mu(\delta_{K_1,K_2})$ & .1229 \\
        $\mu_{K_2}$ & 1.9316 & $\mu(\delta_{K_2,K_3})$ & .1099 \\
        $\mu_{K_3}$ & 2.0415 & $\mu(\delta_{K_3,K_4})$ & .1341 \\
        $\mu_{K_4}$ & 2.1756 & $\mu(\delta_{K_4,K_5})$ & .2223 \\
        $\mu_{K_5}$ & 2.3979 & $\mu(\delta_{K_1,K_5})$ & .5892 \\
        \hline
        \rowcolor{gray!40} $\mu(d_{K_1,K_2})$ & .3321 & \cellcolor{gray!60} $\pi(p_{K_1,K_2}<\alpha')$ & \cellcolor{gray!60} .1030 \\
        \rowcolor{gray!40} $\mu(d_{K_2,K_3})$ & .2424 & \cellcolor{gray!60} $\pi(p_{K_2,K_3}<\alpha')$ & \cellcolor{gray!60} .1030 \\
        \rowcolor{gray!40} $\mu(d_{K_3,K_4})$ & .2451 & \cellcolor{gray!60} $\pi(p_{K_3,K_4}<\alpha')$ & \cellcolor{gray!60} .0515 \\
        \rowcolor{gray!40} $\mu(d_{K_4,K_5})$ & .3461 & \cellcolor{gray!60} $\pi(p_{K_4,K_5}<\alpha')$ & \cellcolor{gray!60} .8754 \\
        \rowcolor{gray!40} $\mu(d_{K_1,K_5})$ & 1.0948 & \cellcolor{gray!60} $\pi(p_{K_1,K_5}<\alpha')$ & \cellcolor{gray!60} 23.0690 \\
        \hline
    \end{tabular}
    \vspace{0.3cm}
    \captionsetup{justification=centering}
    \caption*{\textit{Table n°4: Comparison of mean activations between categorical clusters from GPT4 clustering prompt on tokens' embeddings (layer 1).}}
\end{table}

\begin{figure}[H]
    \centering
    \includegraphics[width=0.6\textwidth]{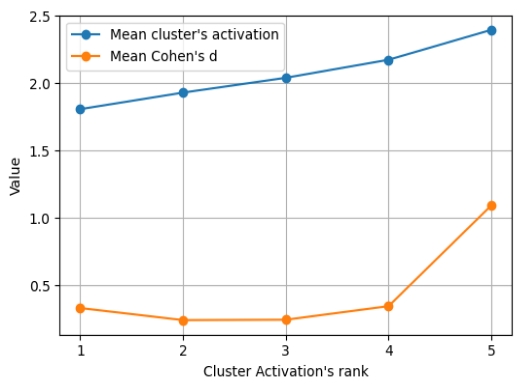}
    \captionsetup{justification=centering, font=small}
    \caption*{\textit{Graphe n°4 : Comparison of mean activations between categorical clusters from GPT4 clustering prompt (layer 1).}}
\end{figure}

From these various initial "top‐down" investigations, a dual trend consistently emerges regarding the existence of a potential synthetic intra‐neuronal attention mechanism that enables a neuron to identify and locate, within the artificial thought category it represents, specific categorical segments determined based on given activation segments:
\begin{itemize}
    \item Activation indifferentiation among categorical clusters corresponding to lower average activation values.
    \item Relative activation differentiation among categorical clusters associated with one or more higher average activation values.
\end{itemize}

\subsection{Activation Interleaving of Categorical Clusters}

We remain within the framework of our \textit{top-down} investigation into the possible existence of a synthetic mechanism of conceptualization and intra-neuronal attention. This mechanism would enable each neuron to identify and differentiate specific categorical segments within its associated thought category, based on particular activation zones. However, we now adopt a significantly different methodological perspective. Here, we examine the extent to which the categorical clusters obtained (using the previous method involving a GPT-4 clustering prompt) form distinct segments in terms of their activation levels. In other words, we investigate whether activation segments specific to categorical clusters overlap (i.e., contain tokens from multiple categorical clusters) or remain distinct (i.e., the activation segment of each categorical cluster contains only tokens from that cluster).

This is operationalized as follows:
\begin{itemize}
    \item Let \( x_{\min}(K_i) \) be the lowest activation value among tokens in categorical cluster \( K_i \), and \( x_{\max}(K_i) \) the highest activation value in the same cluster.
    \item Let \( n(K_i) \) be the number of tokens (from the full set of five clusters \( K_1, K_2, K_3, K_4, K_5 \)) whose activation values fall within the range \( [x_{\min}(K_i), x_{\max}(K_i)] \), and \( m(K_i) \) the number of tokens belonging to cluster \( K_i \).
    \item Let \( N \) be the total number of clustered tokens (100, or fewer when GPT-4 was unable to cluster all tokens).
    \item For a given categorical cluster \( K_i \), there is no overlap if \( n(K_i) = m(K_i) \).
\end{itemize}

For each neuron in layers 0 and 1 of GPT-2XL, we analyzed this operationalization using contingency tables structured as follows and performed an inferential \( \chi^2 \) test. We included only cases where clusters had a theoretical count strictly greater than 5 (to meet the conditions for applying \( \chi^2 \)) and an observed count also strictly greater than 5.

\begin{table}[H]
    \centering
    \renewcommand{\arraystretch}{1.3}
    \begin{tabular}{|c|c|c|}
        \hline
        & \multicolumn{2}{c|}{$K_i$} \\
        \hline
        Observed & $n(K_i)$ & $N - n(K_i)$ \\
        \hline
        Expected & $m(K_i)$ & $N - m(K_i)$ \\
        \hline
        Risk ratio & \multicolumn{2}{c|}{$\frac{n(K_i)}{m(K_i)}$} \\
        \hline
    \end{tabular}
\end{table}

Tables n°5 and n°6, which respectively include 2316 eligible neurons from layer 0 and 1942 eligible neurons from layer 1, reveal strong activation interleaving. The mean risk ratios (\(\mu_{\rho}\)) are consistently high, ranging between 4.5 and 5, indicating a substantial interleaving effect: on average, an activation segment associated with a given categorical cluster contains 4 to 5 times more tokens than the number of tokens explicitly assigned to that cluster. This phenomenon is statistically significant, with very high percentages (\(\pi(p(\chi^2) < .05)\)) of cases where the observed and expected token distributions are markedly different for each cluster.

\begin{table}[H]
    \centering
    \renewcommand{\arraystretch}{1.3}
    \begin{tabular}{|c|c|c|}
        \hline
        N\textsubscript{neuron} & \multicolumn{2}{c|}{2316} \\
        \hline
        & $\mu_{\rho}$ & 5.1103 \\
        \cline{2-3}
        \multirow{-2}{*}{$K_1$} & $\pi(p(\chi^2)<.05)$ & 100 \\
        \hline
        & $\mu_{\rho}$ & 4.9323 \\
        \cline{2-3}
        \multirow{-2}{*}{$K_2$} & $\pi(p(\chi^2)<.05)$ & 100 \\
        \hline
        & $\mu_{\rho}$ & 4.8721 \\
        \cline{2-3}
        \multirow{-2}{*}{$K_3$} & $\pi(p(\chi^2)<.05)$ & 100 \\
        \hline
        & $\mu_{\rho}$ & 4.9705 \\
        \cline{2-3}
        \multirow{-2}{*}{$K_4$} & $\pi(p(\chi^2)<.05)$ & 100 \\
        \hline
        & $\mu_{\rho}$ & 5.0677 \\
        \cline{2-3}
        \multirow{-2}{*}{$K_5$} & $\pi(p(\chi^2)<.05)$ & 99.8705 \\
        \hline
    \end{tabular}
    \captionsetup{justification=centering}
    \caption*{\textit{Table n°5 : Percentages of interleaving activations of categorical core-token clusters (Layer 0).}}
\end{table}

\begin{table}[H]
    \centering
    \renewcommand{\arraystretch}{1.3}
    \begin{tabular}{|c|c|c|}
        \hline
       N\textsubscript{neuron} & \multicolumn{2}{c|}{1942} \\
        \hline
        & $\mu_{\rho}$ & 4.6122 \\
        \cline{2-3}
        \multirow{-2}{*}{$K_1$} & $\pi(p(\chi^2)<.05)$ & 99.9485 \\
        \hline
        & $\mu_{\rho}$ & 4.6212 \\
        \cline{2-3}
        \multirow{-2}{*}{$K_2$} & $\pi(p(\chi^2)<.05)$ & 100 \\
        \hline
        & $\mu_{\rho}$ & 4.6998 \\
        \cline{2-3}
        \multirow{-2}{*}{$K_3$} & $\pi(p(\chi^2)<.05)$ & 99.9485 \\
        \hline
        & $\mu_{\rho}$ & 4.6624 \\
        \cline{2-3}
        \multirow{-2}{*}{$K_4$} & $\pi(p(\chi^2)<.05)$ & 99.9485 \\
        \hline
        & $\mu_{\rho}$ & 4.5496 \\
        \cline{2-3}
        \multirow{-2}{*}{$K_5$} & $\pi(p(\chi^2)<.05)$ & 99.6395 \\
        \hline
    \end{tabular}
    \captionsetup{justification=centering}
    \caption*{\textit{Table n°6 : Percentages of interleaving activations of categorical core-token clusters (Layer 1).}}
\end{table}

The interleaving of activation segments defined by categorical clusters suggests that specific categorical segments are not strictly associated with well-defined activation segments. This result aligns with the \textit{activation indifferentiation} (among categorical clusters with lower mean activation values) highlighted in the previous section. 

Moreover, it does not appear to contradict, at this stage, the \textit{relative activation differentiation} (between categorical clusters with higher mean activation values) mentioned earlier. Indeed, this result does not indicate whether the "excess" tokens (i.e., those interleaved across activation levels) in clusters \( K_5 \) predominantly originate from all categorical clusters (which might contradict the activation differentiation phenomenon) or primarily from clusters \( K_4 \) with fewer from \( K_1 \) (which would be consistent with the activation differentiation phenomenon).

\subsection{Categorical Homogeneity of Activation Clusters}

We now examine the results of our second \textit{bottom-up} study, in which we first isolated specific activation segments and then investigated their respective levels of categorical homogeneity. In other words, we assessed whether tokens from a given activation segment tend to be categorically similar. This analysis directly serves our central research question: does a synthetic process of conceptualization and intra-neuronal attention exist, enabling each neuron to identify and isolate specific categorical segments within the artificial thought category it carries, based on determined activation segments?

From a methodological perspective, categorical proximity was assessed using cosine similarity based on GPT-2XL embeddings. As previously stated, the use of GPT-2XL embeddings is a specific operational choice, which does not represent an absolute semantic ontology but rather one particular way—among others—to measure categorical proximity within a given semantic space. Consequently, our interpretations must be contextualized within this specific semantic contingency.

More precisely, our methodological approach was as follows. For each of the 6,400 neurons in the first two layers of GPT-2XL, we divided the activation space into four activation segments, forming four groups (\( G_1, G_2, G_3, G_4 \)) of tokens, ordered by their mean activation levels. For each of these four groups, we determined its internal semantic homogeneity (\(\cos_{Gi} \)) by computing the mean pairwise cosine similarity of all its tokens. We then calculated an index:

\[
d = \cos_{Gi} - Q_3(\cos_{100})
\]

where \( Q_3(\cos_{100}) \) is the third quartile of the pairwise cosine similarities among the 100 \textit{core-tokens} of the neuron. This index \( d \) thus expresses the extent to which tokens in group \( G_i \) are among the most semantically similar (relative to the overall semantic proximities of tokens within the neuron). Specifically, the more negative \( d \) is, the less semantically similar the concerned tokens are within their neuron.

Initially, we operationalized the segmentation of the activation space into four activation segments using quartiles, producing four groups each containing 25\% of the 100 \textit{core-tokens} for a given neuron. This approach ensured activation clusters were homogeneous in terms of token count. Table n°7 presents the results obtained for the 6,400 neurons in layer 0.

First, we observe that the mean cosine similarities (\(\mu(\cos_{Gi})\)) within the four groups are relatively low, ranging from .34 to .39. Additionally, the percentages (\(\pi(\mu(\cos_{Gi} - Q_3(\cos_{100}) < 0)\)) of cases where mean cosine similarities fall below the third quartile of all pairwise similarities are very high (all above 97\%), indicating that the mean cosine similarities of the \( G_i \) groups are generally not among the highest for each neuron. These percentages are associated with extremely low and thus highly significant inferential probabilities (\( p(\chi^2) \)), assuming a theoretical uniform distribution. These findings suggest a low categorical homogeneity of activation clusters.

A second noteworthy result is that the mean cosine similarities (\(\mu(\cos_{Gi}\)) increase, albeit moderately but systematically, from \( G_1 \) (the least activated tokens, .3485) to \( G_4 \) (the most activated tokens, .3933). Correspondingly, the distances \( \mu(\cos_{Gi}) - Q_3(\cos_{100}) \) tend to decrease slightly in absolute value, as do their associated negativity percentages (\(\pi(\mu(\cos_{Gi}) - Q_3(\cos_{100}) < 0)\)). This suggests a mild but consistent trend of increasing categorical homogeneity in activation clusters for higher activation values. Graph 5 visually synthesizes these results.

\begin{table}[H]
    \centering
    \renewcommand{\arraystretch}{1.3}
    \begin{tabular}{|c|c|}
        \hline
        N\textsubscript{neuron} & 6400 \\
        \hline
        $\mu(\text{cos}_{G_1})$ & .3485 \\
        \hline
        $\mu(\text{cos}_{G_1}) - Q_3(\text{cos}_{100})$ & -.1267 \\
        \hline
        $\pi(\mu(\text{cos}_{G_1}) - Q_3(\text{cos}_{100}) < 0)$ & 99.9531 \\
        \hline
        $p(\chi^2)$ & .0000 \\
        \hline
        $\mu(\text{cos}_{G_2})$ & .3538 \\
        \hline
        $\mu(\text{cos}_{G_2}) - Q_3(\text{cos}_{100})$ & -.1213 \\
        \hline
        $\pi(\mu(\text{cos}_{G_2}) - Q_3(\text{cos}_{100}) < 0)$ & 99.9844 \\
        \hline
        $p(\chi^2)$ & .0000 \\
        \hline
        $\mu(\text{cos}_{G_3})$ & .3644 \\
        \hline
        $\mu(\text{cos}_{G_3}) - Q_3(\text{cos}_{100})$ & -.1108 \\
        \hline
        $\pi(\mu(\text{cos}_{G_3}) - Q_3(\text{cos}_{100}) < 0)$ & 99.9688 \\
        \hline
        $p(\chi^2)$ & .0000 \\
        \hline
        $\mu(\text{cos}_{G_4})$ & .3933 \\
        \hline
        $\mu(\text{cos}_{G_4}) - Q_3(\text{cos}_{100})$ & -.0819 \\
        \hline
        $\pi(\mu(\text{cos}_{G_4}) - Q_3(\text{cos}_{100}) < 0)$ & 97.6250 \\
        \hline
        $p(\chi^2)$ & .0000 \\
        \hline
    \end{tabular}
    \captionsetup{justification=centering}
    \caption*{\textit{Table n°7 : Average cosine similarities of activation clusters by quartiles (Layer 0).}}
\end{table}

\begin{figure}[H]
    \centering
    \includegraphics[width=0.6\textwidth]{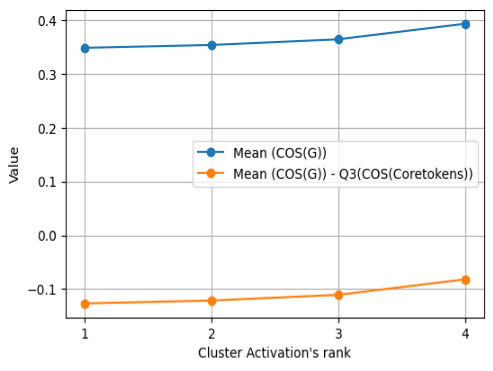}
    \captionsetup{justification=centering, font=small}
    \caption*{\textit{Graph n°5 : Average cosine similarities of activation clusters by quartiles (Layer 0).}}
\end{figure}

Table n°8 and its associated summary graph n°6 exhibit exactly the same type of results for the 6,400 neurons in layer 1. Once again, we observe low \(\mu(\cos_{G_1})\) values, but with a relative progressive increase as activation levels rise, as well as a gradual relative decrease in the gap between \(\mu(\cos_{G_1})\) and \( Q_3(\cos_{100}) \).

\begin{table}[H]
    \centering
    \renewcommand{\arraystretch}{1.3}
    \begin{tabular}{|c|c|}
        \hline
        N\textsubscript{neuron} & 6400 \\
        \hline
        $\mu(\text{cos}_{G_1})$ & .3753 \\
        \hline
        $\mu(\text{cos}_{G_1}) - Q_3(\text{cos}_{100})$ & -.1235 \\
        \hline
        $\pi(\mu(\text{cos}_{G_1}) - Q_3(\text{cos}_{100}) < 0)$ & 99.9219 \\
        \hline
        $p(\chi^2)$ & .0000 \\
        \hline
        $\mu(\text{cos}_{G_2})$ & .3829 \\
        \hline
        $\mu(\text{cos}_{G_2}) - Q_3(\text{cos}_{100})$ & -.1159 \\
        \hline
        $\pi(\mu(\text{cos}_{G_2}) - Q_3(\text{cos}_{100}) < 0)$ & 99.9375 \\
        \hline
        $p(\chi^2)$ & .0000 \\
        \hline
        $\mu(\text{cos}_{G_3})$ & .3956 \\
        \hline
        $\mu(\text{cos}_{G_3}) - Q_3(\text{cos}_{100})$ & -.1033 \\
        \hline
        $\pi(\mu(\text{cos}_{G_3}) - Q_3(\text{cos}_{100}) < 0)$ & 99.9219 \\
        \hline
        $p(\chi^2)$ & .0000 \\
        \hline
        $\mu(\text{cos}_{G_4})$ & .4261 \\
        \hline
        $\mu(\text{cos}_{G_4}) - Q_3(\text{cos}_{100})$ & -.0728 \\
        \hline
        $\pi(\mu(\text{cos}_{G_4}) - Q_3(\text{cos}_{100}) < 0)$ & 96.6563 \\
        \hline
        $p(\chi^2)$ & .0000 \\
        \hline
    \end{tabular}
    \captionsetup{justification=centering}
    \caption*{\textit{Table n°8 : Average cosine similarities of activation clusters by quartiles (Layer 1).}}
\end{table}

\begin{figure}[H]
    \centering
    \includegraphics[width=0.6\textwidth]{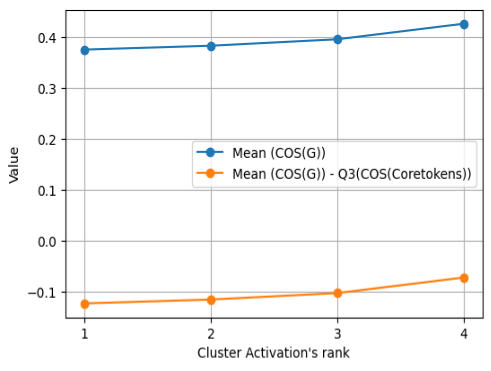}
    \captionsetup{justification=centering, font=small}
    \caption*{\textit{Graph n°6 : Average cosine similarities of activation clusters by quartiles (Layer 1).}}
\end{figure}

In a second phase, and to further diversify our methodology, we applied our \textit{bottom-up} approach by segmenting the activation space into four activation segments, this time using hierarchical clustering on the activation values of the 100 \textit{core-tokens} for each neuron. This method has the advantage of producing activation clusters that are more homogeneous in terms of their mean activation values. This added value is particularly relevant given that our analysis focuses on differentiating categorical proximities based on activation zones. Hierarchical clustering is particularly suited for this task, as it tends to identify activation segments that are more distinct from each other (maximizing inter-cluster variance) while ensuring greater internal homogeneity within each cluster (minimizing intra-cluster variance).

Table n°9 and its corresponding summary graph n°7, covering 6,400 neurons from layer 0, as well as Table n°10 and its related summary graph n°8, covering 6,400 neurons from layer 1, consistently reveal the same synthetic phenomena observed previously. However, in this case, there is a stronger effect of increasing categorical homogeneity in activation clusters for higher activation values.

\begin{table}[H]
    \centering
    \renewcommand{\arraystretch}{1.3}
    \begin{tabular}{|c|c|}
        \hline
        N\textsubscript{neuron} & 6400 \\
        \hline
        $\mu(\text{cos}_{G_1})$ & .3888 \\
        \hline
        $\mu(\text{cos}_{G_1}) - Q_3(\text{cos}_{100})$ & -.0863 \\
        \hline
        $\pi(\mu(\text{cos}_{G_1}) - Q_3(\text{cos}_{100}) < 0)$ & 73.5938 \\
        \hline
        $p(\chi^2)$ & .0000 \\
        \hline
        $\mu(\text{cos}_{G_2})$ & .4318 \\
        \hline
        $\mu(\text{cos}_{G_2}) - Q_3(\text{cos}_{100})$ & -.0433 \\
        \hline
        $\pi(\mu(\text{cos}_{G_2}) - Q_3(\text{cos}_{100}) < 0)$ & 65.4219 \\
        \hline
        $p(\chi^2)$ & .002 \\
        \hline
        $\mu(\text{cos}_{G_3})$ & .4494 \\
        \hline
        $\mu(\text{cos}_{G_3}) - Q_3(\text{cos}_{100})$ & -.0258 \\
        \hline
        $\pi(\mu(\text{cos}_{G_3}) - Q_3(\text{cos}_{100}) < 0)$ & 59.4844 \\
        \hline
        $p(\chi^2)$ & .0578 \\
        \hline
        $\mu(\text{cos}_{G_4})$ & .4665 \\
        \hline
        $\mu(\text{cos}_{G_4}) - Q_3(\text{cos}_{100})$ & -.0087 \\
        \hline
        $\pi(\mu(\text{cos}_{G_4}) - Q_3(\text{cos}_{100}) < 0)$ & 50.5625 \\
        \hline
        $p(\chi^2)$ & .9104 \\
        \hline
    \end{tabular}
    \captionsetup{justification=centering}
    \caption*{\textit{Table n°9 : Average cosine similarities of activation clusters by hierarchical classification (Layer 0).}}
\end{table}

\begin{figure}[H]
    \centering
    \includegraphics[width=0.6\textwidth]{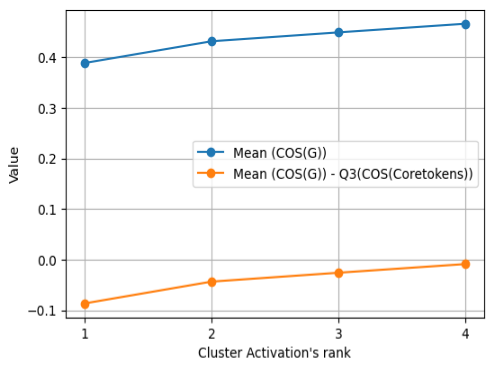}
    \captionsetup{justification=centering, font=small}
    \caption*{\textit{Graph n°7 : Average cosine similarities of activation clusters by hierarchical classification (Layer 0).}}
\end{figure}

\begin{table}[H]
    \centering
    \renewcommand{\arraystretch}{1.3}
    \begin{tabular}{|c|c|}
        \hline
        N\textsubscript{neuron} & 6400 \\
        \hline
        $\mu(\text{cos}_{G_1})$ & .4444 \\
        \hline
        $\mu(\text{cos}_{G_1}) - Q_3(\text{cos}_{100})$ & -.0544 \\
        \hline
        $\pi(\mu(\text{cos}_{G_1}) - Q_3(\text{cos}_{100}) < 0)$ & 66.4688 \\
        \hline
        $p(\chi^2)$ & .001 \\
        \hline
        $\mu(\text{cos}_{G_2})$ & .4769 \\
        \hline
        $\mu(\text{cos}_{G_2}) - Q_3(\text{cos}_{100})$ & -.0219 \\
        \hline
        $\pi(\mu(\text{cos}_{G_2}) - Q_3(\text{cos}_{100}) < 0)$ & 60.5625 \\
        \hline
        $p(\chi^2)$ & .0346 \\
        \hline
        $\mu(\text{cos}_{G_3})$ & .4899 \\
        \hline
        $\mu(\text{cos}_{G_3}) - Q_3(\text{cos}_{100})$ & -.0090 \\
        \hline
        $\pi(\mu(\text{cos}_{G_3}) - Q_3(\text{cos}_{100}) < 0)$ & 55.1875 \\
        \hline
        $p(\chi^2)$ & .2995 \\
        \hline
        $\mu(\text{cos}_{G_4})$ & .5013 \\
        \hline
        $\mu(\text{cos}_{G_4}) - Q_3(\text{cos}_{100})$ & .0024 \\
        \hline
        $\pi(\mu(\text{cos}_{G_4}) - Q_3(\text{cos}_{100}) < 0)$ & 47.0625 \\
        \hline
        $p(\chi^2)$ & .5569 \\
        \hline
    \end{tabular}
    \captionsetup{justification=centering}
    \caption*{\textit{Table n°10 : Average cosine similarities of activation clusters by hierarchical classification (Layer 1).}}
\end{table}

\begin{figure}[H]
    \centering
    \includegraphics[width=0.6\textwidth]{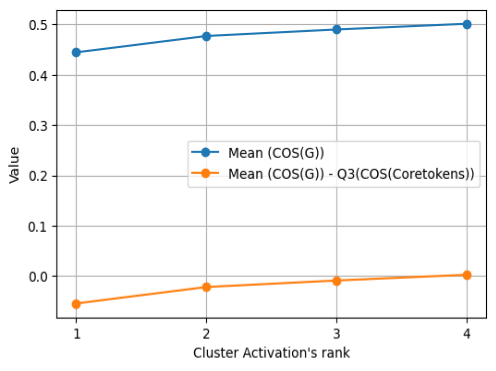}
    \captionsetup{justification=centering, font=small}
    \caption*{\textit{Graph n°8: Average cosine similarities of activation clusters by hierarchical classification (Layer 1).}}
\end{figure}

As part of our final exploration of the existence of a synthetic intra-neuronal attention process, all our results regarding the categorical homogeneity of activation clusters tend to indicate a low categorical homogeneity of activation clusters, accompanied by a gradual positive evolution of this homogeneity as activation values increase.

\section{Discussion and Conclusion}

\subsection{Summary and Interpretation of Results}

The central question of this study was to determine the extent to which the specific activation level of a neuron in response to a token is linked to a categorical value of that token within the neuron's own category. More precisely, in topological terms, we investigated whether a given segment of activation values in a neuron correlates with a distinct categorical segment, thereby enabling the neuron to selectively focus its intra-neuronal attention on a specific categorical segment based on its associated activation range. Put differently, from an epistemological and functional perspective, is the activation space of a neuron compartmentalized into activation zones whose significance and function are to facilitate the attentional detection of specific categorical zones?

Our methodological approach was twofold:

\begin{itemize}
    \item A textit{top-down} approach, where we started from categorical segments to examine their respective mean activation values.
    \item A \textit{bottom-up} approach, where we first isolated activation segments and then investigated their categorical homogeneity. It should be noted, once again, that we made the specific choice of operationalizing categorical homogeneity measurement using only the embedding system of GPT-2XL as our observational framework, acknowledging that other valid approaches could have been pursued.
\end{itemize}

In our \textit{top-down} approach, the first study examined the extent to which significant differences in mean activation values existed between categorical clusters. This revealed two main tendencies regarding the potential existence of a synthetic process of intra-neuronal attention that enables a neuron to identify and localize specific categorical segments within the artificial thought category it represents, based on their respective activation spans:

\begin{itemize}
    \item Activation indifferentiation (i.e., weak differences in mean activation values) among categorical clusters with lower mean activation values.
    \item Relative activation differentiation (i.e., stronger differences in mean activation values) among categorical clusters involving one or more higher mean activation values.
\end{itemize}

Continuing our \textit{top-down} investigation of intra-neuronal attention, we identified activation interleaving among categorical clusters—meaning that given categorical segments are not confined to distinct activation segments but instead overlap. We noted that this finding is compatible with the observed activation indifferentiation (between categorical clusters with lower mean activation values) and, at this stage, does not contradict the observed activation differentiation (between categorical clusters with higher mean activation values).

Finally, in our \textit{bottom-up} study of intra-neuronal attention, our results suggest a low categorical homogeneity of activation clusters, accompanied by a relative but consistent increase in homogeneity as activation values rise.

Overall, these findings are compatible with the existence of an intra-neuronal attention mechanism that can be characterized as follows:

\begin{enumerate}
    \item No direct correlation between activation segmentation and categorical segmentation for strongly activated tokens (since all tokens considered in this study are core-tokens, meaning they are highly activated on average).
    \item A weak but systematic relationship between activation segmentation and categorical segmentation for tokens with the highest activation levels.
\end{enumerate}

In other words, activation appears to play the role of an intra-neuronal attentional vector: it enables a neuron to identify and delimit, within the set of tokens constituting its categorical extension, those tokens specifically associated with the highest activation values. These tokens are relatively more homogeneous from a categorical perspective and therefore constitute a distinct categorical segment of particular interest.

This intra-neuronal attention mechanism, by facilitating attentional focus on only the highest activations, would thus operate an attentional dichotomization: that is, it establishes a qualitative activation threshold (i.e., very high activations) from a quantitative activation continuum, beyond which intra-neuronal selective attention is triggered. This intra-neuronal attention is directly linked to vigilance mechanisms \cite{Motter1999, Chen2024, Hanzal2024, Murray2024} and selective attention processes \cite{Liu2024, Bosker2024, Zhao2024, Lin2024}.

\subsection{Integration of Current and Previous Findings}

The absence of a relationship between activation segmentation and categorical segmentation for tokens with high activation levels is largely compatible with the synthetic process of categorical divergence proposed in one of our previous investigations \cite{Pichat2024a}. Categorical divergence refers to the following two synthetic cognitive phenomena:

\begin{enumerate}
    \item Categorical discontinuity of successive core-tokens in terms of activation levels, meaning that the cosine similarity between successive core-tokens is particularly low.
    \item Categorical inhomogeneity among core-tokens with the same activation levels, indicating that core-tokens sharing similar activation levels are not necessarily the most categorically similar.
\end{enumerate}

This phenomenon of categorical divergence may be linked to the polysemantic nature of neuronal concepts \cite{Olah2020, Fan2023, Bills2023, Bricken2023}, distinguishing it from traditional human categorical approaches \cite{Hornsby2020, Singh2020, Vogel2021, Sanborn2021, Nosofsky2022, Love2022, Poth2023, Roads2024}.

Conversely, the relative correlation we observed between activation and categorical segmentation for tokens with very high activation values aligns with the categorical convergence phenomenon identified in a previous study \cite{Pichat2024b}. Categorical convergence postulates that as the activation levels of successive core-tokens increase, their categorical variability decreases. This aligns with the fundamental characteristic of thought categories being \textit{ad hoc} \cite{Barsalou1995, Glaser2020, Love2022}, meaning they serve a purpose, which, in synthetic categories, might involve minimal alignment with human thought categories and partial convergence toward human-like semantic elements. Furthermore, the tokens specifically involved in this very high activation and categorical convergence process could be interpreted as corresponding to categorical prototypes \cite{Singh2020, Vogel2021}.

In another previous study \cite{Pichat2024c}, we identified three mathematical-cognitive factors influencing categorical segmentation performed by each synthetic neuron:

\begin{itemize}
    \item Categorical priming (effect \( x \)): A token strongly activating a neuron in layer \( n \) has a higher probability of activating a strongly connected neuron in layer \( n+1 \). This categorical priming is conceptually linked to its human counterpart \cite{Anderson1985, Chao2024, Xu2024, HernandezGutierrez2024}.
    \item Inter-categorical attention (effect \( w \)): The stronger the connection between a neuron in layer \( n+1 \) and a neuron in layer \( n \), the higher the probability that a token strongly activated in the first neuron will also be strongly activated in the second.
    \item Categorical phasing (effect \( \Sigma \)): Tokens that simultaneously activate different neurons in layer \( n \) have a higher probability of activating a strongly connected neuron in layer \( n+1 \). This notion of categorical phasing is inspired by similar concepts in cognitive psychology and human neuroscience \cite{Liu2020, Mitchell2021, Kaup2024, Protachevicz2021, CanalesJohnson2021, Ribary2024, Shavikloo2024, Rzechorzek2024}.
\end{itemize}

These three factors drive the determination of the activation value associated with a given token. By mathematical construction of the aggregation function, a strong and combined effect of these three factors on a given token results in a very high activation value for that token. Consequently, this token is positioned within the activation segment of very highly activated tokens for a given neuron. These factors—categorical priming, inter-categorical attention, and categorical phasing—thus enable and guide intra-neuronal attention, which manifests as an \textit{elective attentional focus} on highly activated tokens. These tokens define a specific categorical segment that is particularly relevant for the neuron involved.

In a separate prior study \cite{Pichat2024d}, we identified a synthetic categorical contouring mechanism, which consists of the separation of a categorical form from its background at the neuronal level \cite{Zeki2002, Hock2024, Green2024}. More specifically, categorical contouring is the process by which a specific categorical sub-dimension is extracted \cite{Ghorbani2019, Bhatt2020, Clark2021, Ponomarev2022} from the category vectorized by a precursor neuron (in layer \( n \)) to contribute to the formation of a superordinate category (in layer \( n+1 \)). This contouring is directly driven by the mathematical-cognitive factors mentioned above. The categorical sub-dimension extracted through this process is immediately determined by intra-neuronal attention: by mathematical construction of the neuronal aggregation function, intra-neuronal attention enables the selective focus on a segment of tokens—those with very high activation values—from which the sub-dimensions will statistically be extracted.

Furthermore, we have demonstrated \cite{Pichat2025} that the co-activity of inter-categorical attention and categorical phasing generates a synthetic phenomenon of partial categorical convergence: the categorical sub-dimensions extracted from layer \( n \) categories, when strongly linked to an attentionally connected neuron in layer \( n+1 \), exhibit a tendency toward partial semantic convergence. Categorical convergence is directly influenced by intra-neuronal attention, as this attention mechanism allows for the identification and selection of specific tokens—those with very high activations—through which categorical phasing and, subsequently, categorical convergence take place.

\subsection{Intra-Neuronal Attention and Conceptualization}

As previously discussed, conceptualization \cite{Vergnaud2009, Vergnaud2016} refers to the identification of specific operational characteristics within a class of objects (in this case, tokens) on which a cognitive system must act to ensure appropriate and effective processing. Conceptualization is inherently an attentional phenomenon, as it enables a selective focus on a limited subset of objects or specific attributes of these objects, thereby aligning cognitive activity with their particular properties.

An artificial neuron can be interpreted as a synthetic cognitive agent of conceptualization, whose function is to selectively attend to and filter a subset of tokens from the broader set to which it strongly reacts. This subset constitutes the categorical extension of the \textit{critical concept-in-act} that the neuron is designed to identify selectively. Consequently, the neuron performs an attentional focusing activity on specific token types, selecting and filtering them to optimize the language processing task in which the neuron is engaged.

The paroxysmal concept-in-act thus identified consists of tokens with very high activation levels. These particular tokens, which undergo activation differentiation, tend to semantically converge (categorical homogeneity) toward a characteristic relevant to the neuron's categorical function. The activation differentiation of the categorical segment identified by this synthetic critical concept-in-act is closely linked to the categorical priming effect (effect \( x \)), as it inherently denotes tokens with exceptionally high activation. Effect \( x \), combined with categorical phasing and inter-categorical attention, facilitates categorical contouring as well as categorical convergence.

Thus, intra-neuronal conceptualization and attention, occurring at the level of a neuron in layer \( n \), are the fundamental synthetic processes that subsequently enable the formation of new, more functionally effective thought categories at layer \( n+1 \). These emerging categories further enhance the network's capacity to perform the tasks for which it has been trained.

\section{Conclusion}

In this study, we investigated the extent to which a synthetic process of conceptualization and intra-neuronal attention exists within the perceptron-type neurons of language models. Specifically, we examined whether each neuron can identify and isolate a specific categorical segment within the artificial thought category it represents, based on its activation space. This inquiry was closely tied to the question of whether formal neurons internally exhibit a relative homomorphic relationship between activation segmentation and categorical segmentation, thereby shaping the functional and epistemological significance that can be pragmatically attributed to activation.

Our findings suggest that such a relationship does exist, albeit subtly but systematically, and only at the level of tokens with very high activation values. This intra-neuronal attention mechanism segments, within a given neuron in layer \( n \), an activation zone associated with a specific neuronal concept-in-act. This synthetic concept-in-act serves as the basis for categorical restructuring processes—such as categorical contouring and categorical convergence—which can then be carried out in neurons of layer \( n+1 \). These processes subsequently guide the formation of higher-order categorical abstractions, constituting the thought categories of these superordinate neurons.

It is important to emphasize, once again, that the observed phenomenon of greater categorical homogeneity among tokens with very high activation levels—a central finding in our study of intra-neuronal attention and conceptualization—was operationalized using the input embedding system of GPT-2XL. Additionally, the activation differentiation process we identified was determined using GPT-4o. The advantage of this methodology is that it allows us to study intra-neuronal attention using semantic observation frameworks that are analogous to, or at least relatively aligned with, human thought categories. This seems particularly relevant given that we are investigating how intra-neuronal attention enables a language model to align with human tasks—tasks that inherently require conceptualization and the use of thought categories that are compatible with human cognitive structures.

However, from another epistemological perspective, this approach may be considered biased, self-referential, and anthropomorphic. Indeed, it could be worthwhile to explore the relationship between activation and categorical segmentation within a given neuron using each respective layer's own input embeddings as the semantic observation framework. The added value of this alternative approach—more closely aligned with the unique and layer-specific thought categories and \textit{alien concepts} of each formal neuron—would be its greater fidelity to the categorical structures intrinsic to each neural layer. This alternative methodology might reveal a different phenomenology of intra-neuronal attention and conceptualization, potentially leading to findings such as stronger categorical homogeneity among highly activated tokens or even a broader homomorphic relationship between activation and categorical segmentation, extending beyond the very high activation levels highlighted in our current study.

\section*{Acknowledgments}

The authors thank Madeleine Pichat for her careful review of this article, as well as Chantal Colle for the stimulating projects undertaken with her in the field of artificial intelligence.

\end{document}